\documentclass[a4paper]{article}
\usepackage{graphicx}
\usepackage{twocolceurws}

\title{Convolutional Neural Networks for Sentiment Classification on Business Reviews}

\author{
Andreea Salinca \\ Faculty of Mathematics and Computer Science, University of Bucharest\\ Bucharest, Romania  \\ andreea.salinca@fmi.unibuc.ro}

\institution{}

\begin{document}
\maketitle

\begin{abstract}
Recently Convolutional Neural Networks (CNNs) models have proven remarkable results for text classification and sentiment analysis. In this paper, we present our approach on the task of classifying business reviews using word embeddings on a large-scale dataset provided by Yelp: Yelp 2017 challenge dataset. We compare word-based CNN using several pre-trained word embeddings and end-to-end vector representations for text reviews classification. We conduct several experiments to capture the semantic relationship between business reviews and we use deep learning techniques that prove that the obtained results are competitive with traditional methods. 
\end{abstract}

\section{Introduction}

In recent years, researchers have been investigated the problem of automatic text categorization and sentiment classification - the overall opinion towards the subject matter whether the user review is positive or negative. Sentiment classification is useful in the area of recommender systems and business intelligence applications. 

The effectiveness of applying machine learning techniques in sentiment classification of product or movie reviews is achieved using traditional approaches such as representing text reviews using bag-of-words model and different methods such as Naive Bayes, maximum entropy classification and SVM (Support vector machines) \cite{pang2008opinion, pang2002thumbs, maas2011learning}. Convolutional Neural Networks (CNNs) have achieved remarkable results in the area of sentiment analysis and text classification on large-scale databases \cite{kim2014convolutional, zhang2015sensitivity, johnson2014effective}.

In this article, we conduct an empirical study of a word-based CNNs for sentiment classification using Yelp 2017 challenge dataset \cite{yelpchallenge} that comprises 4.1M user reviews about local business with star rating from 1 to 5. We choose two models for comparison, in which both are word-based CNNs with one or multiple layer of convolution built on top of word vectors by choosing pre-trained or end-to-end learned word representations with different embedding sizes. Previous works report several techniques on sentiment classification results of text reviews using Yelp 2015 challenge dataset \cite{zhang2015character, tang2015document, salinca2015business}.

A series of experiments are made to explore the effect of architecture components on model performance along with the hyperparameters tuning, including filter region size, number of feature maps, and regularization parameters of the proposed convolutional neural networks. We discuss the design decisions for sentiment classification on Yelp 2017 dataset and we offer a comparison between these models and report the obtained accuracy. 

In our work, we aim to identify empirical hyperparameter tuning and practical settings and we inspire from other research conducted by \cite{kim2014convolutional} on a CNNs simple architecture. Furthermore, we also take into consideration some advices from the empirical analysis of CNNs architectures and hyperparameter settings for sentence classification described by \cite{zhang2015sensitivity}. We obtain an accuracy of 95.6\%, via 3-fold cross validation, on Yelp 2017 challenge dataset using word-based CNN along with sentiment-specific word embeddings. 

\section{Prior Work}
Kim et al. present a series of experiments using a simple one layer convolutional neural network built on top of pre-trained word2vec models obtained from an unsupervised neural language model with little parameter tuning for sentiment analysis and sentence classification \cite{kim2014convolutional}.
Zhang et al. offer practical advice by performing an extensive study on the effect of architecture components of CNNs for sentence classification on model performance with results that outperform baseline methods such as SVM or logistic regression \cite{zhang2015sensitivity}.

In \cite{johnson2014effective} it is proven the benefit of word order on topic classification and sentiment classification using CNNs and bag-of-words model in the convolution layer. 

Other approaches use character-level convolutional networks rather than word-based approaches that achieve state of art results for text classification and sentiment analysis on large-scale reviews datasets such as Amazon and Yelp 2015 challenge dataset. For the Yelp polarity dataset, by considering stars 1 and 2 negative, 3 and 4 positive and dropping 5 star reviews, the authors use 560 000 train samples, 38 000 test and 5 000 epochs in training \cite{zhang2015character}. 

A comparison between several models using traditional techniques with several feature extractors: Bag-of-words and TFIDF  (term-frequency inverse-document-frequency), Bag-of-ngrams and TFIDF, Bag-of-means on word embedding – (word2vec) and  TFIDF and a linear classifier - multinomial logistic regression and deep learning techniques: Word-based ConvNets (Convolutional Neural Networks) (one large – 1024 and one small - 256 features sizes having  9 layers deep with 6 convolutional layers and 3 fully-connected layers) and long-short term memory (LSTM) recurrent neural network model is made. The testing errors are reported on all models for Yelp sentiment analysis: 4.36\% is obtained for n-gram traditional approach, word-based CNNs with pre-trained word2vec obtain 4.60\% for the large-featured architecture and 5.56\% for the small-featured architecture. Also, word-based CNNs lookup tables achieve a score of 4.89\% for the large-featured architecture and 5.54\% for the small-featured architecture. The character-level ConvNets model reports an error of 5.89\% for the large-featured architecture and 6.53\% for the small-featured architecture \cite{zhang2015character}.

In \cite{tang2015document} is proposed a convolutional-gated recurrent neural network approach, which encodes relations between sentences and obtains a 67.1\% accuracy on Yelp 2015 dataset (split in training, development and testing sets of 80/10/10) which is compared to a baseline implementation of a convolutional neural network based on Kim work \cite{kim2014convolutional} with an accuracy of 61.5\% for sentiment analysis. On the same dataset, an accuracy of 62.4\% is achieved using a traditional approach with SVM and bigrams.

In prior work, the authors use traditional approaches in the sentiment analysis classification on Yelp 2015 challenge dataset (split in 80\% for training and 20\% for testing and 3-fold cross validation). Linear Support Vector Classification and Stochastic Gradient Descent Classifier report an accuracy of 94.4\% using unigrams and applying preprocessing techniques to extract a set of feature characteristics \cite{salinca2015business}.

\section{Convolutional Neural Network Model}

We model Yelp text reviews using two convolutional architecture approaches. The first model is word-based CNN having an embedding layer in which we tokenize text review sentences to a sentence matrix having rows with word vector representations of each token similar to the approach of Kim et al. \cite{kim2014convolutional}. We will truncate the reviews to a maximum length of 1000 words and we will only consider the top 100 000 most commonly occurring words in the business reviews dataset.

We use both pre-trained word embeddings such as GloVe \cite{karpathy2015deep} – using 100 dimensional embeddings of 400k words computed on a 2014 dump of English Wikipedia, word2vec \cite{mikolov2013efficient} – using 300 dimensional embeddings and fastText \cite{bojanowski2016enriching} – using 300 dimensional embeddings and a vocabulary trained from the reviews dataset using word2vec having 100-dimension word embeddings. Out-of-vocabulary words are randomly initialized by sampling values uniformly from (−0.25, 0.25) and optimized during training.

Next, a convolutional layer with one region sized filters is applied. Filter widths are equal to the dimension of the word vectors  \cite{zhang2015sensitivity}. Then we apply a max-pooling operation on the feature map to compute a fixed-length feature vector and finally a softmax classifier to predict the outputs. During training, we use dropout regularization technique with deep networks where network units are randomly dropped during training \cite{gal2016theoretically}. Also, we aim to minimize the categorical cross-entropy loss. We use a 300 feature maps, 1D convolution window of lengths 2, rectified linear unit (ReLU) activation function and 1-max-pooling of size 2, 0.2 dropout (p) probability.

The second model approach differs from the first approach by using multiple filters for the same region size to learn complementary features from the same regions. We propose 3 filter regions size, having 128 features per filter region, 1D convolution window of length 5, a dropout (d) of 0.5 probability and 1-max-pooling of 35. We compare two different optimizers: Nesterov Adam and RMSprop optimizer\cite{sutskever2013importance}. 

\section{Results And Discussion}
\subsection{Yelp Challenge Dataset}

Yelp 2017 challenge dataset, introduced in the 9th round of Yelp Challenge, comprises user reviews about local businesses in 11 cities across 4 countries with star rating from 1 to 5. The large-scale dataset comprises 4.1M reviews and 947K tips by 1M users for 144K businesses \cite{yelpchallenge}. Yelp 2017 challenge dataset has been updated compared to datasets in previous rounds, such as Yelp 2015 challenge dataset or Yelp 2013 challenge dataset.

We conduct our system evaluation on U.S. cities: Pittsburgh, Charlotte, Urbana-Champaign, Phoenix, Las Vegas, Madison, and Cleveland, having 1 942 339 reviews. For the sentiment analysis classification task, we consider the 1 and 2 star ratings as negative sentiments and 4 and 5 as positive sentiments and we drop the 3 star ratings reviews as the average Yelp review is 3.7. 

Next, we will use two subsets of Yelp 2017 dataset to conduct our experiments, due to computational power constraints.

Our first experiments are done on a smaller subset of Yelp dataset having 8200 training samples, 2000 validation samples and 900 testing samples. We will call this Small Yelp dataset.

Further, we experiment on 82 000 training samples, 20 000 validation samples and 9 000 testing samples. We will call this Big Yelp dataset. 

In the last experiment, we split the large-scale Yelp US dataset into 80\% for training and 20\% for testing.  We use 3-fold cross validation for evaluating different hyperparameters for the deep neural methods. We use accuracy as evaluation metric, which is a standard metric to measure the overall sentiment reviews classification performance \cite{manning1999foundations}.

\subsection{Word Embeddings}
We use several pre-trained models of word embeddings built with an unsupervised learning algorithm for obtaining vector representations of words: GloVe \cite{karpathy2015deep}, word2vec along with pre-trained vectors trained on part of Google News dataset (about 100 billion words). The models contain 100-dimensional vectors for 3 million words and phrases \cite{mikolov2013efficient}.

We use also use fastText pre-trained word vectors for English language which are an extension of word2vec. These vectors in dimension 300 were trained on Wikipedia using the skip-gram model described in \cite{bojanowski2016enriching} with default parameters. 

Moreover, we use in the embedding layer of both proposed CNNs a 100-dimensional word2vec embedding vectors that we have trained using the text reviews in the training dataset. 

\subsection{Experimental Results}
We conduct an empirical exploration on the use of the proposed word-based CNNs architecture for sentiment classification on Yelp business reviews.

In the training phase, we use a batch size of 500 and 3 epochs for the first model approach, and a batch size of 128 and 2 epochs for the second model approach.

We obtain the same accuracy of the classification task of 77.88\% when using 100-dimension and 300-dimension GloVe word embeddings with the first CNNs proposed having 300 features maps and a convolution of window of length 5 on the Small Yelp dataset. 

\begin{figure}[ht]
\begin{center}
\includegraphics[height=5cm]{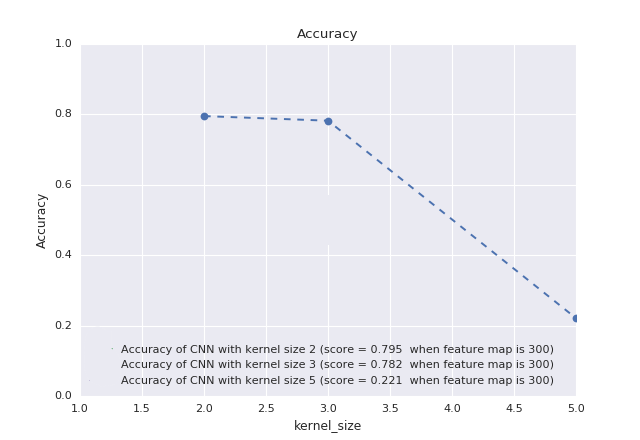}
\caption{CNN accuracy for different kernel sizes when feature map is 300}
\label{accuracyCNN}
\end{center}
\end{figure}

We study the effect of filter kernel size of the convolution when using only one region size on the model accuracy shown in Fig. \ref{accuracyCNN}. We set number of feature maps for this region size to 300 and consider region sizes of 2, 3 and 5 and compute the means of 3-fold CV for each. We observe that using a smaller region size the CNNs performs better, obtaining an accuracy of 79,5\% (window of 2 words) rather than using a larger region size (window size of 5) and obtaining 22,1\%.

The word embeddings used in the embedding layer of our CNNs have successfully captured the semantic relations among entities in the unstructured text review. For the Big Yelp dataset using the first CNN model approach with 300 features map, with a region size of 2, a dropout probability of 0.2 and Nesterov Adam optimizer we obtain a score of 89.59\% in the sentiment classification.

Furthermore, we conduct our study on the second model approach of the word-based CNN having 3 filter regions size, 128 features per filter region, 1D convolution window of length 5, a dropout (d) of 0.5 probability and 1-max-pooling of size 35 along with Nesterov Adam optimizer. 

In Table 1 we report results achieved using the second model approach along with pre-trained GloVe with 100 dimension, word2vec, fastText word embeddings and vocabulary trained from the reviews dataset using word2vec of word embeddings with size of 100. For both pre-trained word2vec and fastText embeddings we choose 300-dimensional word vectors.

We find that the choice of vector input representation has an impact of the performance of the sentiment meaning. On the Small Yelp dataset we report a significand difference of 11.52\% between the highest score using pre-trained GloVe embeddings and self-built dictionary using word2vec model.

However, on the Big Yelp dataset we report a difference of 0.81\% between the highest score using pre-trained fastText embeddings and pre-trained word2vec vectors. The relative performance achieved using the second CNN model approach has similar accuracy scores on the Big Yelp dataset, regardless of the input embeddings (Table 1). We can observe that the scale of the dataset has an impact on the overall performance in the sentiment classification task.

\begin{table}[ht]
\caption{Accuracy results on Yelp reviews dataset.}
\bigskip
\centering
\resizebox{\columnwidth}{!}{%
\begin{tabular}{r|llll}
\multicolumn{1}{r}{Dataset}
& \multicolumn{1}{l}{Model CNN}
& \multicolumn{1}{l}{Embed.dimension}
& \multicolumn{1}{l}{train} 
& \multicolumn{1}{l}{test} \\ \hline
Small Yelp reviews  & Pre-trained GloVe  & 100 & 89.65\% & 87.36\% \\
Small Yelp reviews & Pre-trained word2vec  & 300 & 91.25\% & 90.41\% \\
Small Yelp reviews & Pre-trained fastText  & 300 & 89.90\% & 88.77\% \\
Small Yelp reviews & Word2Vec self-dictionary  & 100 & 79.45\% & 78.89\% \\ \hline
Big Yelp reviews & Pre-trained GloVe  & 100 & 94.46\% & 94.54\% \\
Big Yelp reviews & Pre-trained word2vec  & 300 & 93.80\% & 93.92\% \\
Big Yelp reviews & Pre-trained fastText  & 300 & 94.49\% & 94.73\% \\
Big Yelp reviews & Word2Vec self-dictionary  & 100 & 94.45\% & 94.60\% \\
\hline
\end{tabular}%
}
\end{table}

Training is done through stochastic gradient descent over shuffled mini-batches with Nesterov Adam or RMSprop update rule. Nesterov Adam obtains better results than RMSprop \cite{sutskever2013importance} when using the second model approach with the same number of epochs and a dropout of 0.2. The sentiment accuracy computed on the Big Yelp dataset using RMSprop method scored 0.16 less than the accuracy obtained using Nesterov Adam which scored 95.15

The CNN model in the second approach performed better in the text review classification than the first approach due to the differences in the architecture model and the depth of the convolutional network, the filter region size has a large effect on the classifier performance, for a dropout of 0.5 we obtain 94.54\% compared to 95.15\% for a 0.2 dropout.

When we impose a stronger regularization on the model the performance increases: for a dropout of 0.5 we obtain 94.54\% compared to 95.15\% for a 0.2 dropout. A similar remark about dropout regularization is reported in \cite{zhang2015sensitivity}

Prior work offers a baseline CNN configuration implementing the architectural decisions and hyperparameters of \cite{kim2014convolutional} on Yelp 2015 Challenge dataset for sentiment classification of text review \cite{tang2015document}. The authors report an accuracy of 61.5\%, and propose a new method that represents document with convolutional recurrent neural network, which adaptively encodes semantics of sentences and their relations and achieve 67.6\%. Also, using traditional methods such as SVM and bigrams report a score of 62.4\%. 

In \cite{zhang2015character} the authors propose character-level CNNs that achieve an accuracy of 94.11\% for the large-featured architecture and 93.47\% for the small-featured architecture and compare the obtained results to baseline word-based CNNs with pre-trained word2vec that obtain 95.40\% accuracy for a large-featured architecture and 94.44\% for the small-featured architecture. In their experiments the authors drop 5 star reviews, and use 560 000 train samples, 38 000 test samples from Yelp 2015 challenge dataset and 5 000 epochs in training. Traditional methods as n-grams linear classifier report a score of 95.64\% on the subset.

In comparison against traditional models such as bag of words, n-grams and TFIDF variants, the deep learning models - word-based CNNs and the hyperparameters proposed in this paper obtain comparable to the baseline methods \cite{zhang2015character, tang2015document, salinca2015business}. On the Big Yelp dataset, we report an accuracy of 94.73\% using pre-trained fastText vector embeddings and a CNN having 3 filter regions sizes and 128 feature maps. 

Further, we conduct our evaluation on the complete Yelp 2017 challenge dataset. The second CNN model approach proposed in this work yields the best performance on Yelp 2017 challenge dataset in terms of accuracy. We obtain an accuracy of 95.6\% using 3-fold cross validation.

\section{Conclusions And Future Work}
In the present work, we have described a series of experiments with word-based convolutional neural networks. We introduce two neural network models approaches with different architectural size and several word vector representations. We conduct an empirical study on effect of hyperparameters on the overall performance in the sentiment classification task.

In the experimental results, we find that the size of the dataset has an important effect on the system performance in training and evaluation, a better accuracy score is obtained using the second CNN model approach on the Big Yelp dataset compared to the results obtained on Small Yelp dataset. Furthermore, when evaluating the second model approach on the large scale 2017 Yelp Dataset, we achieve an accuracy score of 95.6\% using 3-fold cross validation.

The models proposed in this article show good ability for understanding natural language and predicting users’ sentiments. We see that our results are comparable and sometimes overcome the ones in the literature for the task of classifying business reviews using Yelp 2017 challenge dataset  \cite{zhang2015character, tang2015document, salinca2015business}. 

In future work, we can explore Bayesian optimization frameworks for hyperparameters ranges rather than a grid search approach. Also, we can conduct other experiments using Recursive Neural Network (RNN) with the Long Short Term Memory (LSTM) architecture \cite{graves2012supervised} for sentiment categorization of Yelp user text reviews.

\bibliographystyle{alpha}
\bibliography{CNN_salinca}

\end{document}